\PassOptionsToPackage{usenames,dvipsnames,table,hyperref}{xcolor}
\documentclass[11pt,a4paper]{article}
\usepackage[]{emnlp2017}
\usepackage{times}
\usepackage{latexsym}
\usepackage{amssymb}
\usepackage{times}
\usepackage{latexsym}
\usepackage{amsmath}
\usepackage{multirow}
\usepackage{url}
\usepackage{amsfonts}
\usepackage{tabularx}
\usepackage{booktabs}
\usepackage{rotating}
\usepackage{subfig}
\usepackage{array}
\usepackage{booktabs}
\usepackage{amsfonts}
\usepackage{enumitem}
\usepackage{tabularx}
\usepackage{makecell}
\usepackage{xcolor}
\usepackage[toc,page]{appendix}
\PassOptionsToPackage{table}{xcolor}
\usepackage{times}
\usepackage{url}
\usepackage{latexsym}
\usepackage{amsmath}
\usepackage{graphics}
\usepackage{graphicx}
\usepackage{color}
\usepackage{stmaryrd } 
 \usepackage{array,multirow,multicol}
\usepackage{tabularx}
\usepackage{makecell}
\usepackage{booktabs} 
\usepackage{lipsum}
\usepackage{amsmath}
\usepackage{algorithm}
\usepackage[noend]{algpseudocode}
\usepackage{mathtools}
\usepackage{enumitem} 
\usepackage{amsfonts}
\usepackage[noend]{algpseudocode}
\usepackage{xcolor}
\usepackage{dblfloatfix} 
\usepackage{float}
\usepackage{mathtools}
\usepackage{bm}
\usepackage{xcolor}
\usepackage{amsmath}

\newcommand{\methoda}{\textsc{CSpikes}}

\newcommand{\methodc}{\textsc{CGraph}}

\let\emptyset\varnothing

\newcommand{\specialcell}[2][c]{%
  \begin{tabular}[#1]{@{}l@{}}#2\end{tabular}}

\def\BState{\State\hskip-\ALG@thistlm}

\DeclareMathOperator*{\argmax}{arg\,max}

\usepackage{etoolbox}
\preto\align{\par\nobreak\normalsize\noindent}
\expandafter\preto\csname align*\endcsname{\par\nobreak\normalsize\noindent}

\emnlpfinalcopy

\title{
Detecting and Explaining Causes From Text For a Time Series Event
}

\author{
Dongyeop Kang, Varun Gangal, Ang Lu, Zheng Chen, Eduard Hovy\\
Language Technology Institute\\
Carnegie Mellon University\\
{\tt \{dongyeok,vgangal,alu1,zhengc1,hovy\}@cs.cmu.edu}
}

\date{}

\begin{document}
\maketitle

\begin{abstract}
Explaining underlying causes or effects about events is a challenging but valuable task.
We define a novel problem of generating explanations of a time series event by (1) searching cause and effect relationships of the time series with textual data and (2) constructing a connecting chain between them to generate an explanation.
To detect causal features from text, we propose a novel method based on the Granger causality of time series between features extracted from text such as N-grams, topics, sentiments, and their composition.
The generation of the sequence of causal entities requires a commonsense causative knowledge base with efficient reasoning. 
To ensure good interpretability and appropriate lexical usage we combine symbolic and neural representations, using a neural reasoning algorithm trained on commonsense causal tuples to predict the next cause step.
Our quantitative and human analysis show empirical evidence that our method successfully extracts meaningful causality relationships between time series with textual features and generates appropriate explanation between them.
\end{abstract}


\section{Introduction}

Producing true causal explanations requires deep understanding of the domain.  
This is beyond the capabilities of modern AI.  
However, it is possible to collect large amounts of causally related events, and, given powerful enough representational variability,  
to construct cause-effect chains by selecting individual pairs appropriately and linking them together.    
Our hypothesis is that chains composed of locally coherent pairs can suggest overall causation.  

In this paper, we view \textit{causality} as (commonsense) cause-effect expressions that occur frequently in online text such as news articles or tweets. For example, ``\textit{greenhouse gases causes global warming}" is a sentence that provides an `atomic' link that can be used in a larger chain.
By connecting such causal facts in a sequence, the result can be regarded as a \textit{causal explanation} between the two ends of the sequence 
(see Table~\ref{tab:exchains} for examples).

\noindent This paper makes the following contributions: 
\begin{itemize}[leftmargin=*,noitemsep,topsep=0pt]
\item we define the problem of causal explanation generation,
\item we detect causal features of a time series event  (\methoda) using Granger~\cite{granger1988some} method with features extracted from text such as N-grams, topics, sentiments, and their composition, 
\item we produce a large graph called \methodc~of local cause-effect units derived from text and develop a method to produce causal explanations by selecting and linking appropriate units, using neural representations to enable unit matching and chaining. 
\end{itemize}

\begin{figure}[t]
\small
\centering
{
{\includegraphics[width=.89\linewidth]{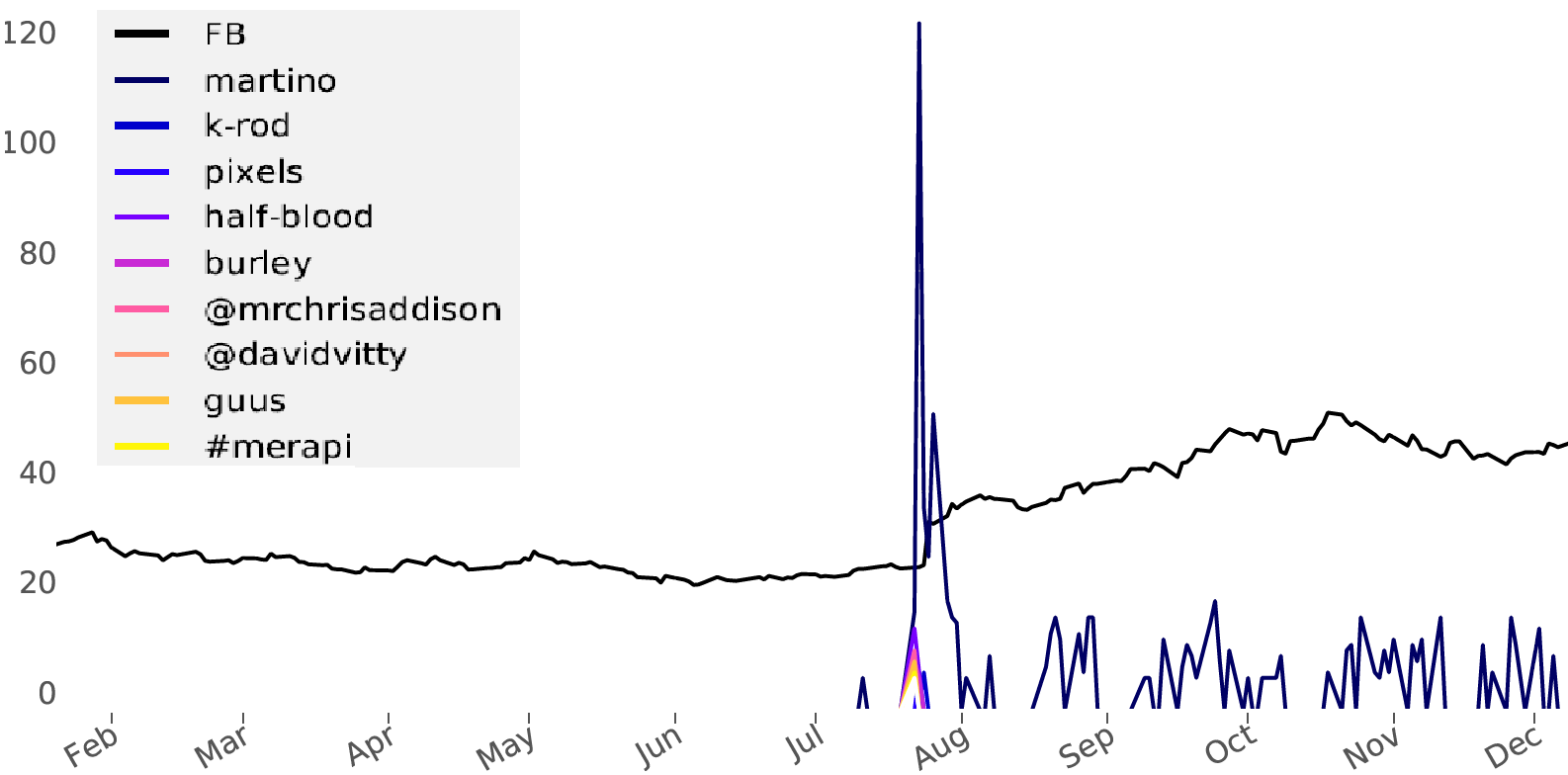}}
}
\caption{\label{fig:example}
Example of causal features for Facebook's stock change in 2013.
The causal features (e.g., \textit{martino}, \textit{k-rod}) rise before the Facebook's rapid stock rise in August.
}
\end{figure}

The problem of causal explanation generation arises for systems that seek to determine causal factors for events of interest automatically.
For given time series events such as companies' stock market prices, our system called \methoda~detects events that are deemed causally related by time series analysis using Granger Causality regression~\cite{granger1988some}.  
We consider a large amount of text and tweets related to each company, and produces for each company time series of values for hundreds of thousands of word n-grams, topic labels, sentiment values, etc.  
Figure~\ref{fig:example} shows an example of causal features that temporally causes Facebook's stock rise in August.

\begin{table}[t]
\centering
\small
\caption{\label{tab:exchains}Examples of generated causal explanation between some temporal causes and target companies' stock prices.
}
\begin{tabularx}{\columnwidth}{@{}X@{}}
\toprule
\textbf{\color{Sepia}party} $\xmapsto[]{cut}$ budget\_cuts $\xmapsto[]{lower}$ budget\_bill $\xmapsto[]{decreas}$ republicans $\xmapsto[]{caus}$ obama $\xmapsto[]{lead to}$ facebook\_polls $\xmapsto[]{caus}$ \textbf{\color{BlueViolet}facebook's stock} $\downarrow$\\\hline
\bottomrule
\end{tabularx}
\end{table}

However, it is difficult to understand how the statistically verified factors actually cause the changes, and whether there is a latent causal structure relating the two. 
This paper addresses the challenge of finding such latent causal structures, in the form of \textit{causal explanations} that connect the given cause-effect pair.
Table~\ref{tab:exchains} shows example causal explanation that our system found between \textit{party} and \textit{Facebook's stock fall ($\downarrow$)}. 

To construct a general causal graph, we extract all potential causal expressions from a large corpus of text. We refer to this graph as \methodc. We use FrameNet~\cite{baker1998berkeley} semantics to provide various causative expressions (verbs, relations, and patterns), 
which we apply to a resource of $183,253,995$ sentences of text and tweets.
These expressions are considerably richer than previous rule-based patterns~\cite{riaz2013toward,kozareva2012cause}.
\methodc~ contains 5,025,636 causal edges. 

Our experiment demonstrates that our causality detection algorithm outperforms other baseline methods for forecasting future time series values. Also, we tested the neural reasoner on the inference generation task using the BLEU score.
Additionally, our human evaluation shows the relative effectiveness of neural reasoners in generating appropriate lexicons in explanations.


\section{\methoda: Temporal Causality Detection from Textual Features}
\label{sec:method}

The objective of our model is, given a target time series $y$, to find the best set of textual features $F = \{f_1, ..., f_k\} \subseteq X$, that maximizes sum of causality over the features on $y$, where $X$ is the set of all features. Note that each feature is itself a time series:
\begin{equation}\label{objective}
   \argmax_{F} \mathbf{C}{ (y, \Phi(X, y)) }
\end{equation}
where $\mathbf{C}(y,x)$ is a causality value function between $y$ and $x$, and $\Phi$ is a linear composition function of features $f$.
$\Phi$ needs target time series $y$ as well because of our graph based feature selection algorithm described in the next sections.

We first introduce the basic principles of Granger causality in Section~\ref{subsec:granger}.  Section~\ref{subsec:feature} describes how to extract good source features $F = \{f_1, ..., f_k\}$ from text. Section~\ref{subsec:causality} describes the causality function $\mathbf{C}$ and the feature composition function $\Phi$.

\subsection{Granger Causality}\label{subsec:granger}

The essential assumption behind Granger causality is that a cause must occur before its effect, and can be used to predict the effect.
Granger showed that given a target time series $y$ (effect) and a source time series $x$ (cause), \textit{forecasting} future target value $y_t$ with both past target and past source time series $E(y_t | y_{<t}, x_{<t})$ is significantly powerful than with only past target time series $E(y_t | y_{<t})$ (plain auto-regression), if $x$ and $y$ are indeed a cause-effect pair.
First, we learn the parameters $\alpha$ and $\beta$ to maximize the prediction expectation:
\begin{align}\label{granger}
    &E(y_t | y_{<t}, x_{t-l})  = \sum_{j=1}^{m} \alpha_j y_{t-j} + \sum_{i=1}^{n} \beta_i x_{t-i} 
\end{align}
where $i$ and $j$ are size of lags in the past observation. Given a pair of causes $x$ and a target $y$, if $\beta$ has magnitude significantly higher than zero (according to a confidence threshold), we can say that $x$ causes $y$.


\subsection{Feature Extraction from Text}\label{subsec:feature}

Extracting meaningful features is a key component to detect causality.
For example, to predict future trend of presidential election poll of \textit{Donald Trump}, we need to consider his past poll data as well as people's reaction about his pledges such as \textit{Immigration}, \textit{Syria} etc.
To extract such ``good'' features crawled from on-line media data, we propose three different types of features: $F_{words}$, $F_{topic}$, and $F_{senti}$.

$F_{words}$ is time series of N-gram words that reflect popularity of the word over time in on-line media.
For each word, the number of items (e.g., tweets, blogs and news) that contains the N-gram word is counted to get the day-by-day time series.
For example,  $x^{\small Michael\_Jordan} = [ 12,51,..]$ is a time series for a bi-gram word \textit{Michael Jordan}.
We filter out stationary words by using simple measures to estimate how dynamically the time series of each word changes over time. Some of the simple measures include Shannon entropy, mean, standard deviation, maximum slope, and number of rise and fall peaks. 

$F_{topic}$ is time series of latent topics with respect to the target time series.
The latent topic is a group of semantically similar words as identified by a standard topic clustering method such as LDA~\cite{blei2003latent}.
To obtain temporal trend of the latent topics, we choose the top ten frequent words in each topic and count their occurrence in the text to get the day-by-day time series.
For example, $x^{healthcare}$ means how popular the topic \textit{healthcare} that consists of \textit{insurance}, \textit{obamacare} etc, is through time.

$F_{senti}$ is time series of sentiments (positive or negative) for each topic.
The top ten frequent words in each topic are used as the keywords, and tweets, blogs and news that contain at least one of these keywords are chosen to calculate the sentiment score.
The day-by-day sentiment series are then obtained by counting positive and negative words using OpinionFinder~\cite{wilson2005recognizing}, and normalized by the total number of the items that day.

\subsection{Temporal Causality Detection}\label{subsec:causality}
We define a causality function $\mathbf{C}$ for calculating causality score between target time series $y$ and source time series $x$.
The causality function $\mathbf{C}$ uses Granger causality~\cite{granger1988some} by fitting the two time series with a Vector AutoRegressive model with exogenous variables (VARX)~\cite{hamilton1994time}: $y_t =  \alpha y_{t-l} + \beta x_{t-l} + \epsilon_t$
where $\epsilon_t$ is a white Gaussian random vector at time $t$ and $l$ is a lag term.
In our problem, the number of source time series $x$ is not single so the prediction happens in the $k$ multi-variate features $X=(f_1, ... f_k)$ so:
\begin{align}
y_t &  =  \alpha y_{t-l} + \bm{\beta} (f_{1,t-l} + ... +  f_{k,t-l}) + \epsilon_t
\end{align}
where $\bm{\alpha}$ and $\bm{\beta}$ is the coefficient matrix of the target $y$ and source $X$ time series respectively, and $\epsilon$ is a residual (prediction error) for each time series.
$\bm{\beta}$ means contributions of each lagged feature $f_{k,t-l}$ to the predicted value $y_t$.
If the variance of $\bm{\beta_k}$ is reduced by the inclusion of the feature terms $f_{k,t-l} \in X $, then it is said that $f_{k,t-l}$ Granger-causes $y$.

Our causality function $\mathbf{C}$ is then $\mathbf{C}(y, f, l) =\Delta(\beta_{y,f,l})$ where $\Delta$ is change of variance by the feature $f$ with lag $l$.
The total Granger causality of target $y$ is computed by summing the change of variance over all lags and all features:
\begin{align}\label{eq:causality}
\mathbf{C}(y, X) =\sum_{k,l} \mathbf{C}(y, f_k, l) 
\end{align}

We compose best set of features $\Phi$ by choosing top $k$ features with highest causality scores for each target $y$.
In practice, due to large amount of computation for pairwise Granger calculation, we make a bipartite graph between features and targets, and address two practical problems: \textit{noisiness} and \textit{hidden edges}.
We filter out noisy edges based on TFIDF and fill out missing values using non-negative matrix factorization (NMF)~\cite{hoyer2004non}.

\begin{table*}[h!]
\centering
\small
\caption{\label{tab:causalgraph} Example (relation, cause, effect) tuples in different categories (manually labeled): \textit{general}, \textit{company}, \textit{country}, and \textit{people}. FrameNet labels related to causation are listed inside parentheses. The number of distinct relation types are 892.}
\resizebox{\textwidth}{!}{%
\begin{tabular}{@{}r|r||r|l@{}}
\toprule
&\textbf{Relation} &  \multicolumn{2}{c}{\textbf{Cause $\mapsto$ Effect $\qquad$}}  \\\hline
\midrule
\parbox[t]{1mm}{\multirow{3}{*}{\rotatebox[origin=c]{90}{\tiny{General}}}} &
causes (Causation) & the virus (Cause) &	aids (Effect) \\
&cause (Causation) & greenhouse gases (Cause) & global warming (Effect)\\
&forced (Causation) & the reality of world war ii (Cause) & the cancellation of the olympics (Effect)\\
\midrule
\parbox[t]{1mm}{\multirow{3}{*}{\rotatebox[origin=c]{90}{\tiny{Company}}}} &heats (Cause\_temperature\_change) & microsoft vague on windows (Item) & legislation battle (Agent) \\
&promotes (Cause\_change\_of\_position\_on\_a\_scale) & chrome (Item) & google (Agent)\\
&makes (Causation) &  twitter (Cause) & love people you 've never met facebook (Effect)\\
\midrule
\parbox[t]{1mm}{\multirow{3}{*}{\rotatebox[origin=c]{90}{\tiny{Country}}}}
&developing (Cause\_to\_make\_progress) & north korea (Agent) & nuclear weapons (Project)\\
&improve (Cause\_to\_make\_progress) & china (Agent) & its human rights record (Project)\\
&forced (Causation) & war with china (Cause) & the japanese to admit , in july 1938 (Effect)\\
\midrule
\parbox[t]{1mm}{\multirow{3}{*}{\rotatebox[origin=c]{90}{\tiny{People}}}}
&attracts (Cause\_motion) & obama (Agent) & more educated voters (Theme)\\
&draws (Cause\_motion) & on america 's economic brains (Goal) & barack obama (Theme) \\
&made (Causation) & michael jordan (Cause) & about \$ 33 million (Effect)\\
\bottomrule
\end{tabular}
}
\end{table*}

\section{\methodc~Construction}\label{sec:graph}

Formally, given source $x$ and target $y$ events that are causally related in time series, if we could find a sequence of cause-effect pairs $(x \mapsto e_1)$, $(e_1 \mapsto e_2)$,  ... $(e_t \mapsto y)$, then $e_1 \mapsto e_2, ... \mapsto e_t$ might be a good causal explanation between $x$ and $y$.
Section~\ref{sec:graph} and \ref{sec:reasoning} describe how to bridge the causal gap between given events ($x$, $y$) by (1) constructing a large general cause-effect graph (\methodc) from text, (2) linking the given events to their equivalent entities in the causal graph by finding the internal paths ($x \mapsto e_1, ... e_t \mapsto y$) as causal explanations, using neural algorithms.

\methodc~is a knowledge base graph where edges are directed and causally related between entities.
To address less representational variability of rule based methods~\cite{girju2003automatic,blanco2008causal,sharp2016creating} in the causal graph construction, we used FrameNet~\cite{baker1998berkeley} semantics.
Using a semantic parser such as SEMAFOR~\cite{chen2010semafor} that produces a FrameNet style analysis of semantic predicate-argument structures, we could obtain lexical tuples of causation in the sentence. 
Since our goal is to collect only causal relations, we extract total 36 causation related frames\footnote{Causation, Cause\_change, Causation\_scenario, Cause\_ benefit\_or\_detriment, Cause\_bodily\_experience, etc.} from the parsed sentences. 

\begin{table}[h]
\centering
\small
\caption{\label{tab:graphstats} Number of sentences parsed, number of entities and tuples, and number of edges (\textit{KB-KB}, \textit{KBcross}) expanded by Freebase in \methodc. }
\resizebox{\columnwidth}{!}{%
\begin{tabular}{c|c|c|c|c}
\toprule
\# Sentences & \# Entities & \# Tuples  & \# \textit{KB-KB} & \# \textit{KBcross}\\ 
\midrule
183,253,995 & 5,623,924 & 5,025,636 & 470,250 & 151,752\\
\bottomrule
\end{tabular}
}
\end{table}

To generate meaningful explanations, high coverage of the knowledge is necessary.
We collect six years of tweets and NYT news articles from 1989 to 2007 (See Experiment section for details). 
In total, our corpus has 1.5 billion tweets and 11 million sentences from news articles. 
The Table~\ref{tab:graphstats} has the number of sentences processed and number of entities, relations, and tuples in the final \methodc.

Since the tuples extracted from text are very noisy~\footnote{SEMAFOR has around $62\%$ of accuracy on held-out set.}, we constructed a large causal graph by linking the tuples with string match and filter out the noisy nodes and edges based on some graph statistics. 
We filter out nodes with very high degree that are mostly stop-words or auto-generated sentences. 
Too long or short sentences are also filtered out.
Table~\ref{tab:causalgraph} shows the (case, relation, effect) tuples with manually annotated categories such as \textit{General}, \textit{Company}, \textit{Country}, and \textit{People}.

\section{Causal Reasoning}\label{sec:reasoning}

To generate a causal explanation using \methodc, we need traversing the graph for finding the path between given source and target events.
This section describes how to efficiently traverse the graph by expanding entities with external knowledge base and how to find (or generate) appropriate causal paths to suggest an explanation using symbolic and neural reasoning algorithms.

\subsection{Entity Expansion with Knowledge Base}

A simple choice for traversing a graph are the traditional graph searching algorithms such as Breadth-First Search (BFS). However, the graph searching procedure is likely to be incomplete (\textit{low recall}), because simple string match is insufficient to match an effect to all its related entities, as it misses out in the case where an entity is semantically related but has a lexically different name.



To address the \textit{low recall} problem and generate better explanations, we propose the use of knowledge base to augment our text-based causal graph with real-world semantic knowledge. 
We use Freebase~\cite{freebase} as the external knowledge base for this purpose. 
Among $1.9$ billion edges in original Freebase dump, we collect its first and second hop neighbours for each target events.

While our \methodc~is lexical in nature, Freebase entities appear as identifiers (MIDs). 
For entity linking between two knowledge graphs, we need to annotate Freebase entities with their lexical names by looking at the wiki URLs.
We refer to the edges with freebase expansion as \textit{KB-KB} edges, and link the \textit{KB-KB} with our \methodc~using lexical matching, referring as \textit{KBcross} edges (See Table~\ref{tab:graphstats} for the number of the edges).




\subsection{Symbolic Reasoning}

Simple traversal algorithms such as BFS are infeasible for traversing the \methodc~due to the large number of nodes and edges.
To reduce the search space $k$ in $e_{t} \mapsto \{e_{t+1}^1, ...e_{t+1}^k\}$, we restricted our search by  depth of paths, length of words in entity's name, and edge weight. 

\begin{algorithm}[h]
\caption{ Backward Causal Inference. $y$ is target event, $d$ is depth of BFS, $l$ is lag size, $BFS_{back}$ is Breadth-First search for one depth in backward direction, and $\sum_l\mathbf{C}$ is sum of Granger causality over the lags. \label{alg:inference}}
\begin{algorithmic}[1]
\State $\mathbb{S} \gets {\textit{y}}$,  $d = 0 $ 
\While {($\mathbb{S} = \emptyset$) or ($d > D_{max} $)}
	\State $\{e_{-d}^1, ...e_{-d}^k\} \gets BFS_{back}(\mathbb{S})$
	\State $d = d + 1$, $\mathbb{S} \gets \emptyset $
    \For{\texttt{$j$ in $\{1,...,k\}$}}
    	\If {\textit{$\sum_l\mathbf{C}(y,e_{-d}^j,l) < \epsilon$}} $\mathbb{S} \gets e_{-d}^j $
        \EndIf
    \EndFor
\EndWhile
\end{algorithmic}
\end{algorithm}

For more efficient inference, we propose a backward algorithm that searches potential causes (instead of effects) $\{e_{t}^1, ...e_{t}^k\} \mapsfrom e_{t+1}$ starting from the target node $y = e_{t+1}$ using Breadth-first search (BFS).
It keeps searching backward until the node $e_{i}^j$ has less Granger confident causality with the target node $y$ (See Algorithm~\ref{eq:causality} for causality calculation).  
This is only possible because our system has temporal causality measure between two time series events.
See Algorithm~\ref{alg:inference} for detail.

\subsection{Neural Reasoning}

While symbolic inference is fast and straightforward, the sparsity of edges may make our inference semantically poor.
To address the \textit{lexical sparseness}, we propose a lexically relaxed reasoning using a neural network.

Inspired by recent success on alignment task such as machine translation~\cite{bahdanau2014neural}, our model learns the causal alignment between cause phrase and effect phrase for each type of relation between them.
Rather than traversing the \methodc, our neural reasoner uses \methodc~ as a training resource.
The encoder, a recurrent neural network such as LSTM~\cite{hochreiter1997long}, takes the causal phrase while the decoder, another LSTM, takes the effectual phrase with their relation specific attention.

In original attention model~\cite{bahdanau2014neural}, the contextual vector $c$ is computed by $c_i = a_{ij} * h_j$ where $h_j$ is hidden state of causal sequence at time $j$ and $a_{ij}$ is soft attention weight, trained by feed forward network $a_{ij} = FF (h_j, s_{i-1})$ between input hidden state $h_j$ and output hidden state $s_{i-1}$.
The global attention matrix $a$, however, is easy to mix up all local alignment patterns of each relation. 

\begin{figure}[t]
\centering
\includegraphics[trim=2.2cm 5.5cm 2.1cm 5cm,clip,width=.96\linewidth]{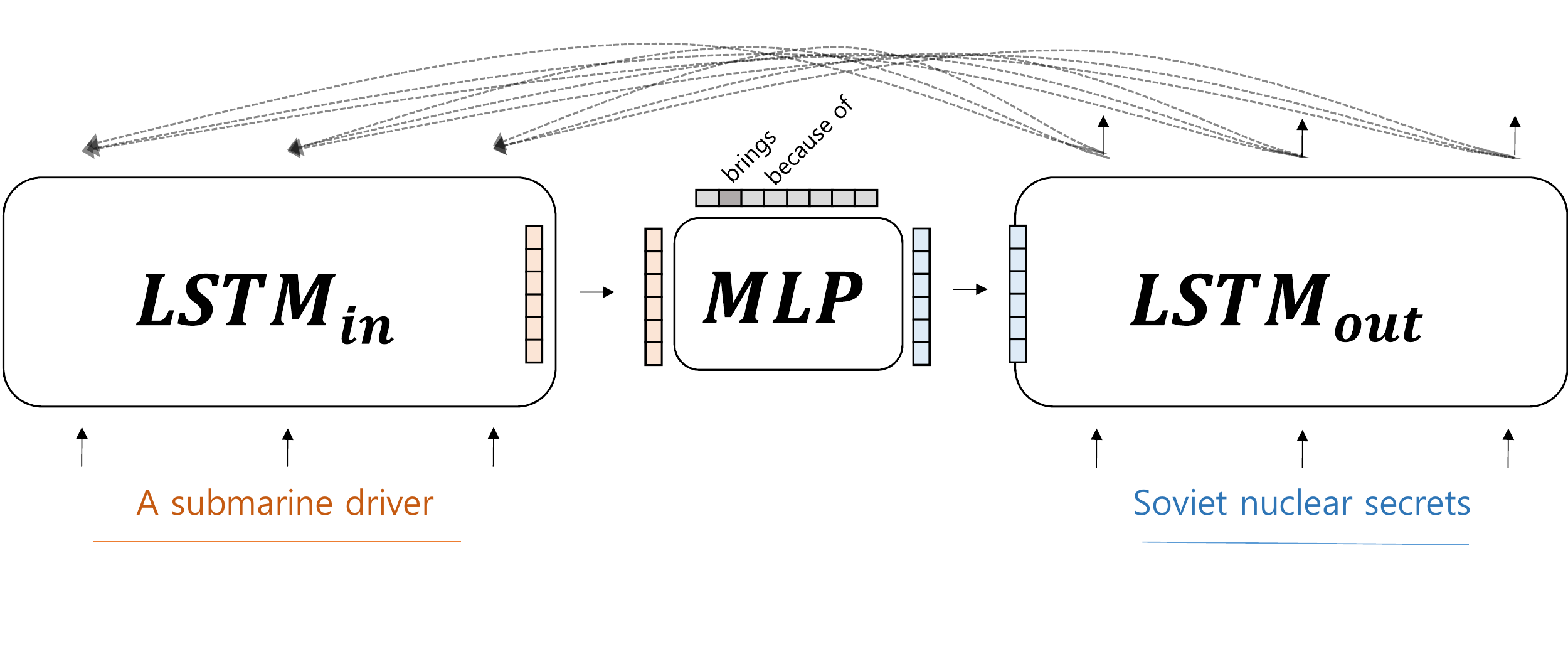}
\caption{\label{fig:proposedmodel} Our neural reasoner. The encoder takes causal phrases and decoder takes effect phrases by learning the causal alignment between them. The MLP layer in the middle takes different types of FrameNet relation and locally attend the cause to the effect w.r.t the relation (e.g., ``because of'', ``led to'', etc). 
}
\end{figure}

For example, a tuple, 
\textit{\small{(north korea (Agent)
$\xmapsto[(Cause\_to\_make\_progress)]{developing}$ nuclear weapons (Project))}}, 
 is different with another tuple,
\textit{\small{(chrome (Item) $\xmapsto[(Cause \_change\_of\_position)]{promotes}$ google (Agent))}} in terms of local type of causality.
To deal with the \textit{local attention}, we decomposed the attention weight $a_{ij}$ by relation specific transformation in feed forward network:
\begin{align*}
&a_{ij} = FF (h_j, s_{i-1}, r)
\end{align*}
where $FF$ has relation specific hidden layer and $r \in R$ is a type of relation in the distinct set of relations $R$ in training corpus  (See Figure~\ref{fig:proposedmodel}).

Since training only with our causal graph may not be rich enough for dealing various lexical variation in text, we use pre-trained word embedding such as word2vec~\cite{mikolov2013distributed} trained on GoogleNews corpus\footnote{https://code.google.com/archive/p/word2vec/} for initialization.
For example, given a cause phrase \textit{weapon equipped}, our model could generate multiple effect phrases with their likelihood: \textit{($\xmapsto[0.54]{result}$war)}, \textit{($\xmapsto[0.12]{force}$army reorganized)}, etc, even though there are no tuples exactly matched in \methodc. 

We trained our neural reasoner in either forward or backward direction.
In prediction, decoder inferences by predicting effect (or cause) phrase in forward (or backward) direction. 
As described in the Algorithm~\ref{alg:inference}, the backward inference continue predicting the previous causal phrases until it has high enough Granger confidence with the target event.

\section{Experiment}\label{sec:experiment}
\textbf{Data}. We collect on-line social media from tweets, news articles, and blogs.
Our Twitter data has one million tweets per day from 2008 to 2013 that are crawled using Twitter's Garden Hose API.
News and Blog dataset have been crawled from 2010 to 2013 using Google's news API.
For target time series, we collect companies' stock prices in NASDAQ and NYSE from 2001 until present for 6,200 companies.
For presidential election polls, we collect polling data of the 2012 presidential election from 6 different websites, including USA Today , Huffington Post, Reuters, etc. 

\begin{table}[t]
\centering
\setlength\tabcolsep{2pt}
\caption{\label{tab:dynamics} Examples of $F_{words}$ with their temporal dynamics: Shannon entropy, mean, standard deviation, slope of peak, and number of peaks.}
\resizebox{\columnwidth}{!}{%
\begin{tabular}{@{}r|cccccc@{}}
\toprule
&  \textbf{entropy}  & \textbf{mean}  & \textbf{STD} & \textbf{max\_slope} & \textbf{\#-peaks}      \\
\midrule
\#lukewilliamss & 0.72 & 22.01 & 18.12 & 6.12 & 31 \\
happy\_thanksgiving &	0.40	& 61.24&	945.95	&3423.75	&414 \\
michael\_jackson	& 0.46	&141.93	&701.97	 &389.19	&585	\\
\bottomrule
\end{tabular}
}
\end{table}

\textbf{Features}. For N-gram word features $F_{word}$,we choose the spiking words based on their temporal dynamics (See Table~\ref{tab:dynamics}).
For example, if a word is too frequent or the time series is too burst, the word should be filtered out because the trend is too general to be an event.
We choose five types of temporal dynamics: Shannon entropy, mean, standard deviation, maximum slope of peak, and number of peaks; 
and delete words that have too low or high entropy, too low mean and deviation, or the number of peaks and its slope is less than a certain threshold.
Also, we filter out words whose frequency is less than five.
From the $1,677,583$ original words, we retain $21,120$ words as final candidates for $F_{words}$ including uni-gram and bi-gram words.

For sentiment $F_{senti}$ and topic $F_{topic}$ features, we choose 50 topics generated for both politicians and companies separately using LDA, and then use top 10 words for each topic to calculate sentiment score for this topic.
Then we can analyze the causality between sentiment series of a specific topic and collected time series.

\textbf{Tasks}. To show validity of causality detector, first we conduct random analysis between target time series and randomly generated time series.
Then, we tested forecasting stock prices and election poll values with or without the detected textual features to check effectiveness of our causal features.
We evaluate our reasoning algorithm for generation ability compared to held-out cause-effect tuples using BLEU metric. 
Then, for some companies' time series, we describe some qualitative result of some interesting causal text features found with Granger causation and explanations generated by our reasoners between the target and the causal features.
We also conducted human evaluation on the explanations.

\subsection{Random Causality Analysis}

\begin{figure}[t]
    \centering
     {
          \subfloat[\footnotesize{$y \xleftarrow[]{lag=3} rf_1, ..., rf_k$ }]{
         \fbox{\includegraphics[clip,trim=2.4cm 0.3cm 0.5cm 1.5cm,width=.84\linewidth, height=60px]{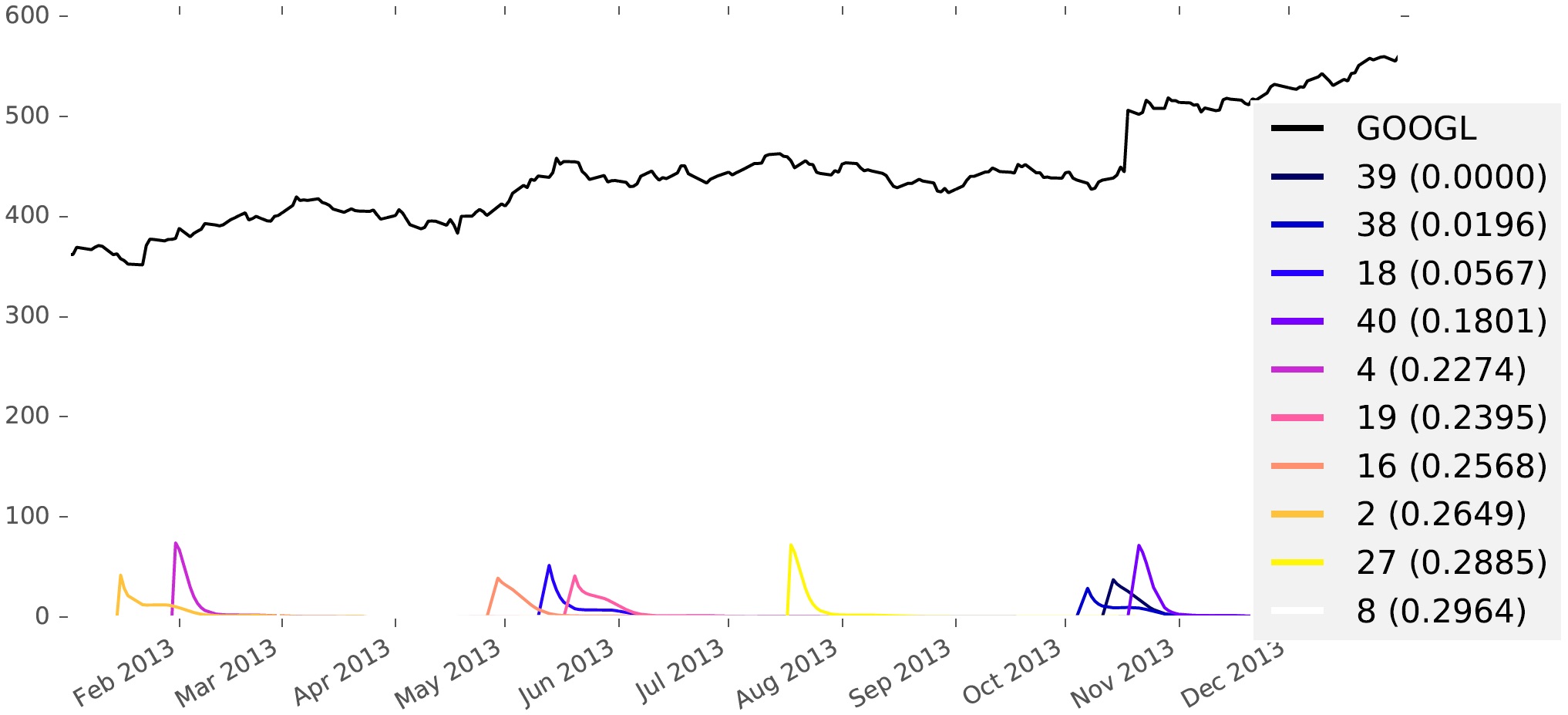}
         }}\\
          \subfloat[\footnotesize{$y \xrightarrow[]{lag=3} rf_1, ..., rf_k$ }]{
         \fbox{\includegraphics[clip,trim=2.4cm 0.3cm 0.5cm 0.7cm,width=.84\linewidth, height=60px]{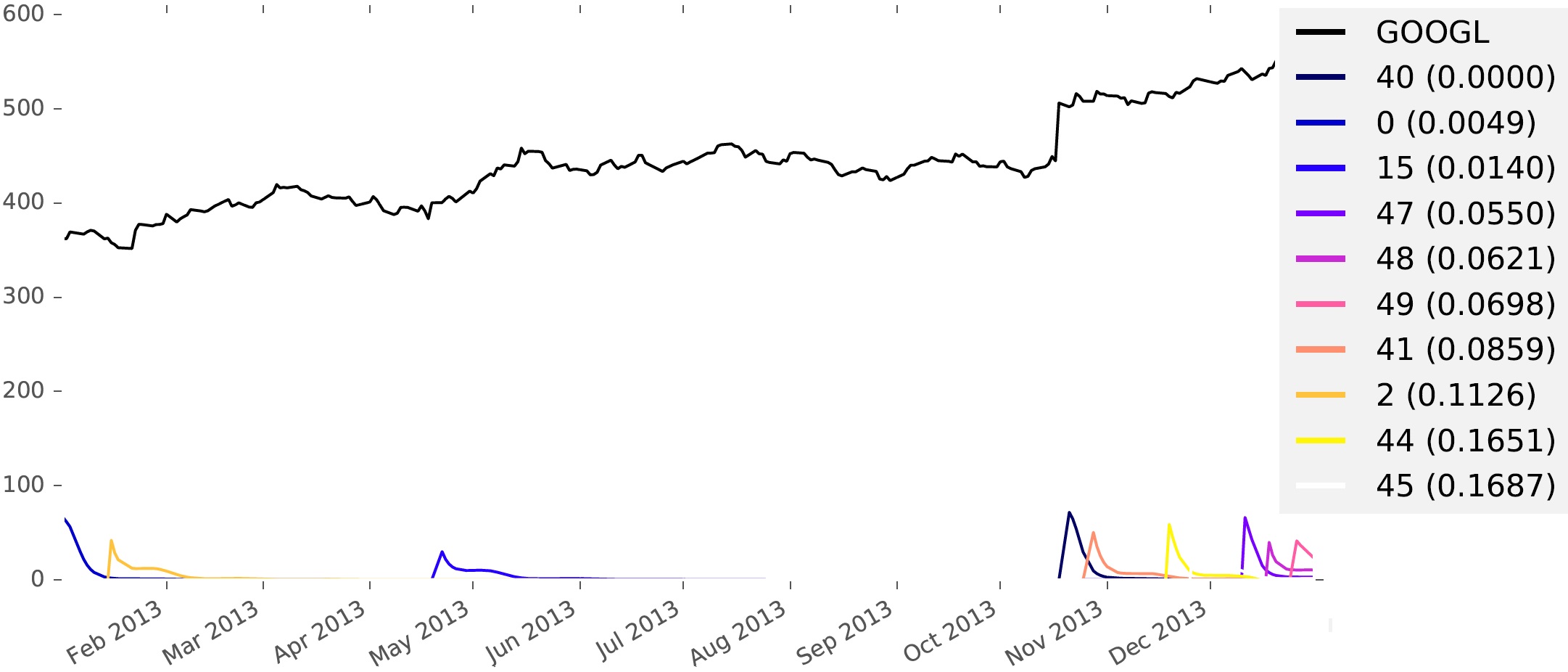}
         }}\\
     }
    \caption{\label{fig:random} Random causality analysis on \textbf{Googles}'s stock price change ($y$) and randomly generated features ($rf$) during 2013-01-01 to 2013-12-31.
    (a) shows how the random features $rf$ cause the target $y$, while (b) shows how the target $y$ causes the random features $rf$ with lag size of 3 days.    
    The color changes according to causality confidence to the target (blue is the strongest, and yellow is the weakest). 
    The target time series has y scale of prices, while random features have y scale of causality degree  $\mathbf{C}(y,rf) \subset [ 0,1 ]$.
    }
\end{figure}

To check whether our causality scoring function $\mathbf{C}$ detects the temporal causality well, we conduct a random analysis between target time series and randomly generated time series (See Figure~\ref{fig:random}).
For Google's stock time series, we regularly move window size of 30 over the time and generate five days of time series with a random peak strength using a SpikeM model~\cite{DBLP:conf/kdd/MatsubaraSPLF12a}\footnote{SpikeM has specific parameters for modeling a time series such as peak strength, length, etc.}.
The color of random time series $rf$ changes from blue to yellow according to causality degree with the target $\mathbf{C}(y,rf)$.
For example, blue is the strongest causality with target time series, while yellow is the weakest.

We observe that the strong causal (blue) features are detected just before (or after) the rapid rise of Google' stock price on middle October in (a) (or in (b)).
With the lag size of three days, we observe that the strength of the random time series gradually decreases as it grows apart from the peak of target event. 
The random analysis shows that our causality function $\mathbf{C}$ appropriately finds cause or effect relation between two time series in regard of their strength and distance.

\subsection{Forecasting with Textual Features}\label{sec:forecasting}
\begin{table}[h]
\footnotesize
\caption{\label{tab:forecasting} Forecasting errors (RMSE) on \textbf{Stock} and \textbf{Poll} data with time series only (\textit{SpikeM} and \textit{LSTM}) and with time series plus text feature (\textit{random}, \textit{words}, \textit{topics}, \textit{sentiment}, and \textit{composition}).}
\centering
\setlength\tabcolsep{2pt}
\begin{tabular}{r|r|cc|ccccc}
\toprule
 \multicolumn{2}{r}{\textit{}} & \multicolumn{2}{c}{\textbf{Time Series}} & \multicolumn{5}{c}{\textbf{Time Series + Text}} \\\hline
 \multicolumn{2}{r}{\textit{Step}} &   SpikeM    & LSTM & $\mathbf{C}_{rand}$ &  $\mathbf{C}_{words}$ & $\mathbf{C}_{topics}$ & $\mathbf{C}_{senti}$ & $\mathbf{C}_{comp}$\\
\midrule
\multirow{3}{2mm}{\rotatebox[origin=c]{90}{\textbf{Stock}}}
&1 & 102.13 & 6.80 & 3.63  & 2.97 & 3.01 & 3.34 & \underline{1.96} \\
&3 & 99.8     & 7.51 & 4.47  & 4.22 & 4.65 & 4.87 & \underline{3.78} \\
&5& 97.99  & 7.79 & 5.32   & \underline{5.25} & 5.44 & 5.95 & 5.28 \\
\hline
\multirow{3}{2mm}{\rotatebox[origin=c]{90}{\textbf{Poll}}}
&1 &10.13 &  1.46    &1.52      & 1.27 & 1.59 & 2.09 & \underline{1.11} \\
&3 & 10.63  & 1.89    & 1.84   &  1.56 & 1.88 & 1.94 & \underline{1.49}\\
&5 & 11.13  & 2.04     & 2.15   & 1.84 & 1.88 & 1.96 &\underline{1.82}\\
\bottomrule
\end{tabular}
\end{table}
We use time series forecasting task as an evaluation metric of whether our textual features are appropriately causing the target time series or not.
Our feature composition function $\Phi$ is used to extract good causal features for forecasting.
We test forecasting on stock price of companies (\textbf{Stock}) and predicting poll value for presidential election (\textbf{Poll}).
For stock data, We collect daily closing stock prices during 2013 for ten IT companies\footnote{Company symbols used: TSLA, MSFT, GOOGL, YHOO, FB, IBM, ORCL, AMZN, AAPL and HPO}. 
For poll data, we choose ten candidate politicians~\footnote{Name of politicians used: Santorum, Romney, Pual, Perry, Obama, Huntsman, Gingrich, Cain, Bachmann} in the period of presidential election in 2012. 

\begin{table}[t]
\centering
\small
\caption{\label{tab:beam} Beam search results in neural reasoning. These examples could be filtered out by graph heuristics before generating final explanation though. }
\begin{tabular}{@{}l@{}|l@{}}
\toprule
Cause$\mapsto$Effect in \methodc & Beam Predictions\\\hline
\midrule
\specialcell{the dollar's \\$\xmapsto[]{caus}$ against the yen} &  \specialcell{$[1]$$\xmapsto[]{caus}$ against the yen\\ $[2]$$\xmapsto[]{caus}$ against the dollar \\ $[3]$$\xmapsto[]{caus}$ against other currencies}  \\\hline
\specialcell{without any exercise \\$\xmapsto[]{caus}$ news article} & \specialcell{$[1]$$\xmapsto[]{lead to}$ a difference \\ $[2]$$\xmapsto[]{caus}$ the risk  \\ $[3]$$\xmapsto[]{make}$ their weight} \\
\bottomrule
\end{tabular}
\end{table}

For each of stock and poll data, the future trend of target is predicted only with target's past time series or with target's past time series and past time series of textual features found by our system.
Forecasting only with target's past time series uses \textit{SpikeM}~\cite{DBLP:conf/kdd/MatsubaraSPLF12a} that models a time series with small number of parameters and simple \textit{LSTM}~\cite{hochreiter1997long,nnet} based time series model.
Forecasting with target and textual features' time series use Vector AutoRegressive model with exogenous variables (VARX)~\cite{hamilton1994time} from different composition function such as $\mathbf{C}_{random}$,  $\mathbf{C}_{words}$, $\mathbf{C}_{topics}$, $\mathbf{C}_{senti}$, and $\mathbf{C}_{composition}$.
Each composition function except $\mathbf{C}_{random}$ uses top ten textual features that causes each target time series.
We also tested LSTM with past time series and textual features but VARX outperforms LSTM. 

Table~\ref{tab:forecasting} shows root mean square error (RMSE) for forecasting with different step size (time steps to predict), different set of features, and different regression algorithms on stock and poll data.
The forecasting error is summation of errors over moving a window (30 days) by 10 days over the period.
Our $\mathbf{C}_{composition}$ method outperforms other time series only models and time series plus text models in both stock and poll data.

\subsection{Generating Causality with Neural Reasoner}
The reasoner needs to predict the next effect phrase (or previous cause phrase) so the model should be evaluated in terms of generation task.
We used the BLEU ~\cite{papineni2002bleu} metric to evaluate the predicted phrases on held out phrases in our \methodc~.
Since our \methodc~ has many edges, there may be many good paths (explanations), possibly making our prediction diverse. To evaluate such diversity in prediction, we used ranking-based BLEU method on the $k$ set of predicted phrases by beam search.
For example, $B@k$ means BLEU scores for generating $k$ number of sentences and $B@kA$ means the average of them.

Table~\ref{tab:beam} shows some examples of our beam search results when $k=3$.
Given a cause phrase, the neural reasoner sometime predicts semantically similar phrases (e.g., \textit{against the yen}, \textit{against the dollar}), while it sometimes predicts very diverse phrases (e.g., \textit{a different}, \textit{the risk}).

Table~\ref{tab:bleu} shows BLEU ranking results with different reasoning algorithms: \textbf{S2S} is a  sequence to sequence learning trained on \methodc~by default, \textbf{S2S+WE} adds word embedding initialization, and \textbf{S2S+REL+WE} adds relation specific attention.
Initializing with pre-trained word embeddings (\textbf{+WE}) helps us improve on prediction.
Our relation specific attention model outperforms the others, indicating that different type of relations have different alignment patterns.

\begin{table}[t]
\centering
\caption{\label{tab:bleu} BLEU ranking.
Additional word representation \textbf{+WE} and relation specific alignment \textbf{+REL} help the model learn the cause and effect generation task especially for diverse patterns.
 }
\begin{tabular}{@{}l|c|c|c}
\toprule
&B@1 & B@3A & B@5A\\
\midrule
\textbf{S2S} & 10.15 & 8.80  & 8.69 \\
\textbf{S2S + WE}  & 11.86 & 10.78  & 10.04  \\
\textbf{S2S + WE + REL} & 12.42 & 12.28 & 11.53 \\
\bottomrule
\end{tabular}
\end{table}

\subsection{Generating Explanation by Connecting}

\begin{table*}[h]
\centering
\caption{\label{tab:explanation} Example causal chains for explaining the rise ($\uparrow$) and fall ($\downarrow$) of companies' stock price. The temporally causal {\color{Sepia}$feature$} and {\color{BlueViolet}$target$} are linked through a sequence of predicted cause-effect tuples by different reasoning algorithms: a symbolic graph traverse algorithm \textit{SYMB} and a neural causality reasoning model \textit{NEUR}.}
\resizebox{\linewidth}{!}{%
\begin{tabular}{@{}c|l@{}}
\midrule
\parbox[t]{1.0mm}{\multirow{3}{*}{\rotatebox[origin=c]{90}{\textit{SYMB}}}}  
& \textbf{\color{Sepia}medals} $\xmapsto[]{match}$ gold\_and\_silver\_medals $\xmapsto[]{swept}$ korea $\xmapsto[]{improving}$ relations  $\xmapsto[]{widened}$ gap $\xmapsto[]{widens}$ \textbf{\color{BlueViolet}facebook} $\uparrow$ \\
& \textbf{\color{Sepia}excess}$\xmapsto[]{match}$excess\_materialism$\xmapsto[]{cause}$people\_make\_films$\xmapsto[]{make}$money $\xmapsto[]{changed}$ twitter $\xmapsto[]{turned} $\textbf{\color{BlueViolet}facebook} $\downarrow$\\
& \textbf{\color{Sepia}clinton}  $\xmapsto[]{match}$president\_clinton $\xmapsto[]{raised}$antitrust\_case $\xmapsto[]{match}$government's\_antitrust\_case\_against\_microsoft $\xmapsto[]{match}$microsoft $\xmapsto[]{beats}$\textbf{\color{BlueViolet}apple} $\downarrow$\\
\hline


\parbox[t]{1.0mm}{\multirow{4}{*}{\rotatebox[origin=c]{90}{\textit{NEUR}}}} 
& \textbf{\color{Sepia}google} $\xmapsto[]{forc}$ microsoft\_to\_buy\_computer\_company\_dell\_announces\_recall\_of\_batteries $\xmapsto[]{cause}$ \textbf{\color{BlueViolet}microsoft} $\uparrow$\\
& \textbf{\color{Sepia}the\_deal} $\xmapsto[]{make}$ money $\xmapsto[]{rais}$ at\_warner\_music\_and\_google\_with\_protest\_videos\_things $\xmapsto[]{caus}$ \textbf{\color{BlueViolet}google} $\downarrow$\\
& \textbf{\color{Sepia}party} $\xmapsto[]{cut}$ budget\_cuts$\xmapsto[]{lower}$ budget\_bill$\xmapsto[]{decreas}$ republicans$\xmapsto[]{caus}$ obama$\xmapsto[]{lead to}$ facebook\_polls$\xmapsto[]{caus}$ \textbf{\color{BlueViolet}facebook} $\downarrow$\\
& \textbf{\color{Sepia}company} $\xmapsto[]{forc}$ to\_stock\_price $\xmapsto[]{lead to}$ investors $\xmapsto[]{increas}$ oracle\_s\_stock $\xmapsto[]{increas}$ \textbf{\color{BlueViolet}oracle} $\uparrow$
\\
\bottomrule
\end{tabular}
}
\end{table*}

Evaluating whether a sequence of phrases is reasonable as an explanation is very challenging task.
Unfortunately, due to lack of quantitative evaluation measures for the task, we conduct a human annotation experiment.

Table~\ref{tab:explanation} shows example causal chains for the rise ($\uparrow$) and fall ($\downarrow$) of companies' stock price, continuously produced by two reasoners: \textit{SYBM} is symbolic reasoner and \textit{NEUR} is neural reasoner. 

\begin{table}[h]
\centering
\caption{\label{tab:eval} Human evaluation on explanation chains generated by symbolic and neural reasoners. }
\begin{tabular}{r|c|c}
\toprule
\textbf{Reasoners} &SYMB   & NEUR\\
\midrule
\textbf{Accuracy (\%)}& 42.5 &  57.5   \\
\bottomrule
\end{tabular}
\end{table}
We also conduct a human assessment on the explanation chains produced by the two reasoners, asking people to choose more convincing explanation chains for each feature-target pair. 
Table~\ref{tab:eval} shows their relative preferences.

\section{Related Work}\label{sec:related}

Prior works on causality detection~\cite{acharya2014causal,websummary,qiu2012granger} in time series data (e.g., gene sequence, stock prices, temperature) mainly use Granger~\cite{granger1988some} ability for predicting future values of a time series using past values of its own and another time series.
\cite{hlavavckova2007causality} studies more theoretical investigation for measuring causal influence in multivariate time series based on the entropy and mutual information estimation.
However, none of them attempts generating explanation on the temporal causality.

Previous works on text causality detection use syntactic patterns such as $X \xmapsto[]{verb} Y$, where the $verb$ is causative~\cite{girju2003automatic,riaz2013toward,kozareva2012cause,do2011minimally} with additional features~\cite{blanco2008causal}.
\cite{kozareva2012cause} extracted cause-effect relations, where the pattern for bootstrapping has a form of $X^* \xmapsto[Z^*]{verb} Y$ from which terms $X^*$ and $Z^*$ was learned.
The syntax based approaches, however, are not robust to semantic variation.

As a part of SemEval~\cite{girju2007semeval}, \cite{mirza2016catena} also uses syntactic causative patterns~\cite{mirza2014analysis} and supervised classifier to achieve the state-of-the-art performance.
Extracting the cause-effect tuples with such syntactic features or temporality~\cite{bethard2008building} would be our next step for better causal graph construction. 

\cite{grivaz2010human} conducts very insightful annotation study of what features are used in human reasoning on causation.
Beyond the linguistic tests and causal chains for explaining causality in our work, other features such as counterfactuality, temporal order, and ontological asymmetry remain as our future direction to study.

Textual entailment also seeks a directional relation between two given text fragments~\cite{dagan2006pascal}.
Recently, \cite{rocktaschel2015reasoning} developed an attention-based neural network method, trained on large annotated pairs of textual entailment, for classifying the types of relations with decomposable attention~\cite{parikh2016decomposable} or sequential tree structure~\cite{chen2016enhancing}.
However, the dataset~\cite{bowman2015large} used for training entailment deals with just three categories, \textit{contradiction}, \textit{neutral}, and \textit{entailment}, and focuses on relatively simple lexical and syntactic transformations~\cite{kolesnyk2016generating}. Our causal explanation generation task is also similar to \textit{future scenario generation}~\cite{hashimoto2014toward,hashimoto2015generating}.
Their scoring function uses heuristic filters and is not robust to lexical variation.


\section{Conclusion}\label{sec:conclusion}
This paper defines the novel task of detecting and explaining causes from text for a time series. 
First, we detect causal features from online text. Then, we construct a large cause-effect graph using FrameNet semantics. By training our relation specific neural network on paths from this graph, our model generates causality with richer lexical variation. We could produce a chain of cause and effect pairs as an explanation which shows some appropriateness. Incorporating aspects such as time, location and other event properties remains a point for future work.
In our following work, we collect a sequence of causal chains verified by domain experts for more solid evaluation of generating explanations.

\bibliography{explanation}

\begin{thebibliography}{37}
\expandafter\ifx\csname natexlab\endcsname\relax\def\natexlab#1{#1}\fi

\bibitem[{nne(2015)}]{nnet}
 2015.
\newblock {Neural network architecture for time series forecasting.}
\newblock \url{https://github.com/hawk31/nnet-ts}.

\bibitem[{Acharya(2014)}]{acharya2014causal}
Saurav Acharya. 2014.
\newblock Causal modeling and prediction over event streams.

\bibitem[{Anand(2014)}]{websummary}
Surya Pratap Singh~Tanwar Anand, Mehndiratta. 2014.
\newblock {Web Metric Summarization using Causal Relationship Graph}.

\bibitem[{Bahdanau et~al.(2014)Bahdanau, Cho, and Bengio}]{bahdanau2014neural}
Dzmitry Bahdanau, Kyunghyun Cho, and Yoshua Bengio. 2014.
\newblock Neural machine translation by jointly learning to align and
  translate.
\newblock \emph{arXiv preprint arXiv:1409.0473}.

\bibitem[{Baker et~al.(1998)Baker, Fillmore, and Lowe}]{baker1998berkeley}
Collin~F Baker, Charles~J Fillmore, and John~B Lowe. 1998.
\newblock The berkeley framenet project.
\newblock In \emph{Proceedings of the 36th Annual Meeting of the Association
  for Computational Linguistics and 17th International Conference on
  Computational Linguistics-Volume 1}, pages 86--90. Association for
  Computational Linguistics.

\bibitem[{Bethard et~al.(2008)Bethard, Corvey, Klingenstein, and
  Martin}]{bethard2008building}
Steven Bethard, William~J Corvey, Sara Klingenstein, and James~H Martin. 2008.
\newblock Building a corpus of temporal-causal structure.
\newblock In \emph{LREC}.

\bibitem[{Blanco et~al.(2008)Blanco, Castell, and Moldovan}]{blanco2008causal}
Eduardo Blanco, Nuria Castell, and Dan~I Moldovan. 2008.
\newblock Causal relation extraction.
\newblock In \emph{LREC}.

\bibitem[{Blei et~al.(2003)Blei, Ng, and Jordan}]{blei2003latent}
David~M Blei, Andrew~Y Ng, and Michael~I Jordan. 2003.
\newblock Latent dirichlet allocation.
\newblock \emph{JMLR}.

\bibitem[{Bowman et~al.(2015)Bowman, Angeli, Potts, and
  Manning}]{bowman2015large}
Samuel~R Bowman, Gabor Angeli, Christopher Potts, and Christopher~D Manning.
  2015.
\newblock A large annotated corpus for learning natural language inference.
\newblock \emph{arXiv preprint arXiv:1508.05326}.

\bibitem[{Chen et~al.(2010)Chen, Schneider, Das, and Smith}]{chen2010semafor}
Desai Chen, Nathan Schneider, Dipanjan Das, and Noah~A Smith. 2010.
\newblock Semafor: Frame argument resolution with log-linear models.
\newblock In \emph{Proceedings of the 5th international workshop on semantic
  evaluation}, pages 264--267. Association for Computational Linguistics.

\bibitem[{Chen et~al.(2016)Chen, Zhu, Ling, Wei, and Jiang}]{chen2016enhancing}
Qian Chen, Xiaodan Zhu, Zhenhua Ling, Si~Wei, and Hui Jiang. 2016.
\newblock Enhancing and combining sequential and tree lstm for natural language
  inference.
\newblock \emph{arXiv preprint arXiv:1609.06038}.

\bibitem[{Dagan et~al.(2006)Dagan, Glickman, and Magnini}]{dagan2006pascal}
Ido Dagan, Oren Glickman, and Bernardo Magnini. 2006.
\newblock The pascal recognising textual entailment challenge.
\newblock In \emph{Machine learning challenges. evaluating predictive
  uncertainty, visual object classification, and recognising tectual
  entailment}, pages 177--190. Springer.

\bibitem[{Do et~al.(2011)Do, Chan, and Roth}]{do2011minimally}
Quang~Xuan Do, Yee~Seng Chan, and Dan Roth. 2011.
\newblock Minimally supervised event causality identification.
\newblock In \emph{Proceedings of the Conference on Empirical Methods in
  Natural Language Processing}, pages 294--303. Association for Computational
  Linguistics.

\bibitem[{Girju(2003)}]{girju2003automatic}
Roxana Girju. 2003.
\newblock Automatic detection of causal relations for question answering.
\newblock In \emph{Proceedings of the ACL 2003 workshop on Multilingual
  summarization and question answering-Volume 12}, pages 76--83. Association
  for Computational Linguistics.

\bibitem[{Girju et~al.(2007)Girju, Nakov, Nastase, Szpakowicz, Turney, and
  Yuret}]{girju2007semeval}
Roxana Girju, Preslav Nakov, Vivi Nastase, Stan Szpakowicz, Peter Turney, and
  Deniz Yuret. 2007.
\newblock Semeval-2007 task 04: Classification of semantic relations between
  nominals.
\newblock In \emph{Proceedings of the 4th International Workshop on Semantic
  Evaluations}, pages 13--18. Association for Computational Linguistics.

\bibitem[{Google(2016)}]{freebase}
Google. 2016.
\newblock {Freebase Data Dumps}.
\newblock \url{https://developers.google.com/freebase/data}.

\bibitem[{Granger(1988)}]{granger1988some}
Clive~WJ Granger. 1988.
\newblock Some recent development in a concept of causality.
\newblock \emph{Journal of econometrics}, 39(1):199--211.

\bibitem[{Grivaz(2010)}]{grivaz2010human}
C{\'e}cile Grivaz. 2010.
\newblock Human judgements on causation in french texts.
\newblock In \emph{LREC}.

\bibitem[{Hamilton(1994)}]{hamilton1994time}
James~Douglas Hamilton. 1994.
\newblock \emph{Time series analysis}, volume~2.
\newblock Princeton university press Princeton.

\bibitem[{Hashimoto et~al.(2015)Hashimoto, Torisawa, Kloetzer, and
  Oh}]{hashimoto2015generating}
Chikara Hashimoto, Kentaro Torisawa, Julien Kloetzer, and Jong-Hoon Oh. 2015.
\newblock Generating event causality hypotheses through semantic relations.
\newblock In \emph{AAAI}, pages 2396--2403.

\bibitem[{Hashimoto et~al.(2014)Hashimoto, Torisawa, Kloetzer, Sano, Varga, Oh,
  and Kidawara}]{hashimoto2014toward}
Chikara Hashimoto, Kentaro Torisawa, Julien Kloetzer, Motoki Sano, Istv{\'a}n
  Varga, Jong-Hoon Oh, and Yutaka Kidawara. 2014.
\newblock Toward future scenario generation: Extracting event causality
  exploiting semantic relation, context, and association features.
\newblock In \emph{ACL (1)}, pages 987--997.

\bibitem[{Hlav{\'a}{\v{c}}kov{\'a}-Schindler
  et~al.(2007)Hlav{\'a}{\v{c}}kov{\'a}-Schindler, Palu{\v{s}}, Vejmelka, and
  Bhattacharya}]{hlavavckova2007causality}
Katerina Hlav{\'a}{\v{c}}kov{\'a}-Schindler, Milan Palu{\v{s}}, Martin
  Vejmelka, and Joydeep Bhattacharya. 2007.
\newblock Causality detection based on information-theoretic approaches in time
  series analysis.
\newblock \emph{Physics Reports}, 441(1):1--46.

\bibitem[{Hochreiter and Schmidhuber(1997)}]{hochreiter1997long}
Sepp Hochreiter and J{\"u}rgen Schmidhuber. 1997.
\newblock Long short-term memory.
\newblock \emph{Neural computation}, 9(8):1735--1780.

\bibitem[{Hoyer(2004)}]{hoyer2004non}
Patrik~O Hoyer. 2004.
\newblock Non-negative matrix factorization with sparseness constraints.
\newblock \emph{Journal of machine learning research}, 5(Nov):1457--1469.

\bibitem[{Kolesnyk et~al.(2016)Kolesnyk, Rockt{\"a}schel, and
  Riedel}]{kolesnyk2016generating}
Vladyslav Kolesnyk, Tim Rockt{\"a}schel, and Sebastian Riedel. 2016.
\newblock Generating natural language inference chains.
\newblock \emph{arXiv preprint arXiv:1606.01404}.

\bibitem[{Kozareva(2012)}]{kozareva2012cause}
Zornitsa Kozareva. 2012.
\newblock Cause-effect relation learning.
\newblock In \emph{Workshop Proceedings of TextGraphs-7 on Graph-based Methods
  for Natural Language Processing}, pages 39--43. Association for Computational
  Linguistics.

\bibitem[{Matsubara et~al.(2012)Matsubara, Sakurai, Prakash, Li, and
  Faloutsos}]{DBLP:conf/kdd/MatsubaraSPLF12a}
Yasuko Matsubara, Yasushi Sakurai, B.~Aditya Prakash, Lei Li, and Christos
  Faloutsos. 2012.
\newblock Rise and fall patterns of information diffusion: model and
  implications.
\newblock In \emph{KDD}, pages 6--14.

\bibitem[{Mikolov and Dean(2013)}]{mikolov2013distributed}
T~Mikolov and J~Dean. 2013.
\newblock Distributed representations of words and phrases and their
  compositionality.
\newblock \emph{Advances in neural information processing systems}.

\bibitem[{Mirza and Tonelli(2014)}]{mirza2014analysis}
Paramita Mirza and Sara Tonelli. 2014.
\newblock An analysis of causality between events and its relation to temporal
  information.
\newblock In \emph{COLING}, pages 2097--2106.

\bibitem[{Mirza and Tonelli(2016)}]{mirza2016catena}
Paramita Mirza and Sara Tonelli. 2016.
\newblock Catena: Causal and temporal relation extraction from natural language
  texts.
\newblock In \emph{The 26th International Conference on Computational
  Linguistics}, pages 64--75.

\bibitem[{Papineni et~al.(2002)Papineni, Roukos, Ward, and
  Zhu}]{papineni2002bleu}
Kishore Papineni, Salim Roukos, Todd Ward, and Wei-Jing Zhu. 2002.
\newblock Bleu: a method for automatic evaluation of machine translation.
\newblock In \emph{Proceedings of the 40th annual meeting on association for
  computational linguistics}, pages 311--318. Association for Computational
  Linguistics.

\bibitem[{Parikh et~al.(2016)Parikh, T{\"a}ckstr{\"o}m, Das, and
  Uszkoreit}]{parikh2016decomposable}
Ankur~P Parikh, Oscar T{\"a}ckstr{\"o}m, Dipanjan Das, and Jakob Uszkoreit.
  2016.
\newblock A decomposable attention model for natural language inference.
\newblock \emph{arXiv preprint arXiv:1606.01933}.

\bibitem[{Qiu et~al.(2012)Qiu, Liu, Subrahmanya, and Li}]{qiu2012granger}
Huida Qiu, Yan Liu, Niranjan~A Subrahmanya, and Weichang Li. 2012.
\newblock Granger causality for time-series anomaly detection.
\newblock In \emph{Data Mining (ICDM), 2012 IEEE 12th International Conference
  on}, pages 1074--1079. IEEE.

\bibitem[{Riaz and Girju(2013)}]{riaz2013toward}
Mehwish Riaz and Roxana Girju. 2013.
\newblock Toward a better understanding of causality between verbal events:
  Extraction and analysis of the causal power of verb-verb associations.
\newblock In \emph{Proceedings of the annual SIGdial Meeting on Discourse and
  Dialogue (SIGDIAL)}. Citeseer.

\bibitem[{Rockt{\"a}schel et~al.(2015)Rockt{\"a}schel, Grefenstette, Hermann,
  Ko{\v{c}}isk{\`y}, and Blunsom}]{rocktaschel2015reasoning}
Tim Rockt{\"a}schel, Edward Grefenstette, Karl~Moritz Hermann, Tom{\'a}{\v{s}}
  Ko{\v{c}}isk{\`y}, and Phil Blunsom. 2015.
\newblock Reasoning about entailment with neural attention.
\newblock \emph{arXiv preprint arXiv:1509.06664}.

\bibitem[{Sharp et~al.(2016)Sharp, Surdeanu, Jansen, Clark, and
  Hammond}]{sharp2016creating}
Rebecca Sharp, Mihai Surdeanu, Peter Jansen, Peter Clark, and Michael Hammond.
  2016.
\newblock Creating causal embeddings for question answering with minimal
  supervision.
\newblock \emph{arXiv preprint arXiv:1609.08097}.

\bibitem[{Wilson et~al.(2005)Wilson, Wiebe, and
  Hoffmann}]{wilson2005recognizing}
Theresa Wilson, Janyce Wiebe, and Paul Hoffmann. 2005.
\newblock Recognizing contextual polarity in phrase-level sentiment analysis.
\newblock In \emph{Proceedings of the conference on human language technology
  and empirical methods in natural language processing}, pages 347--354.
  Association for Computational Linguistics.

\end{thebibliography}
\bibliographystyle{emnlp_natbib}

\end{document}